\documentclass[conference]{IEEEtran}
\IEEEoverridecommandlockouts
\usepackage{cite}
\usepackage{amsmath,amssymb,amsfonts}
\usepackage{algorithmic}
\usepackage{graphicx}
\usepackage{textcomp}
\usepackage{xcolor}
\usepackage{booktabs}
\usepackage{multirow}
\usepackage{xcolor}

\usepackage{graphicx}
\usepackage{multirow}
\usepackage{bbding}
\usepackage{enumitem}
\usepackage[colorlinks,linkcolor=black,anchorcolor=black,citecolor=black]{hyperref}
\def\BibTeX{{\rm B\kern-.05em{\sc i\kern-.025em b}\kern-.08em
    T\kern-.1667em\lower.7ex\hbox{E}\kern-.125emX}}
\begin{document}

% \title{Conference Paper Title*\\
% {\footnotesize \textsuperscript{*}Note: Sub-titles are not captured for https://ieeexplore.ieee.org  and

\title{UV-Mamba: A DCN-Enhanced State Space Model for Urban Village Boundary Identification in High-Resolution Remote Sensing Images \thanks{$^*$Lulin Li and Ben Chen contribute equally. $^\dagger$Corresponding author.}}

\author{\IEEEauthorblockN{Lulin~Li$^{1,*}$, Ben Chen$^{1,*}$, Xuechao Zou$^{2}$, Junliang Xing$^{3}$, Pin Tao$^{1,3,\dagger}$}
\IEEEauthorblockA{$^1$School of Computer Technology and Applications, Qinghai University, Xining, China \\
$^2$School of Computer Science and Technology, Beijing Jiaotong University, Beijing, China \\
$^3$Department of Computer Science  and Technology, Tsinghua University, Beijing, China \\
\{lulinlee, benchen1997\}@163.com, xuechaozou@foxmail.com, \{jlxing, taopin\}@tsinghua.edu.cn }
}

% % \author{\IEEEauthorblockN{Lulin~Li\IEEEauthorrefmark{1}, Homer Simpson\IEEEauthorrefmark{2}, James K
% irk\IEEEauthorrefmark{3}, Montgomery Scott\IEEEauthorrefmark{3} and Eldon Tyrell\IEEEauthorrefmark{4}}
% \IEEEauthorblockA{\IEEEauthorrefmark{1}School of Computer Technology and Applications, Qinghai University, Xining, China \\
% School of Computer Science and Technology, Beijing Jiaotong University, Beijing, China \\
% Department of Computer Science  and Technology, Tsinghua University, Beijing, China \\
% Email: \{lulinlee, benchen1997\}@163.com, xuechaozou@bjtu.edu.cn, \{jlxing, taopin\}@tsinghua.edu.cn }
% }

\maketitle

\begin{abstract}
% Due to the diverse geographical environments, intricate landscapes, and high-density settlements, the automatic identification of urban village boundaries using remote sensing images is a highly challenging task. This paper proposes a novel and efficient neural network model called UV-Mamba for accurate boundary detection in high-resolution remote sensing images. UV-Mamba mitigates the memory loss problem in lengthy sequence modeling, which arises in state space model (SSM) with increasing image size, by incorporating deformable convolutions (DCN). Its architecture utilizes an encoder-decoder framework and includes an encoder with four deformable state space augmentation blocks for efficient multi-level semantic extraction and a decoder to integrate the extracted semantic information. We conducted experiments on the Beijing and Xi'an datasets, and the results show that UV-Mamba achieves state-of-the-art performance. Specifically, our model achieves 73.3\% and 78.1\% IoU on the Beijing and Xi'an datasets, respectively, representing improvements of 1.2\% and 3.4\% IoU over the previous best model while also being 6x faster in inference speed and 40x smaller in parameter count. Source code and pre-trained models are available at \url{https://github.com/Devin-Egber/UV-Mamba}.
Due to the diverse geographical environments, intricate landscapes, and high-density settlements, the automatic identification of urban village boundaries using remote sensing images remains a highly challenging task. This paper proposes a novel and efficient neural network model called UV-Mamba for accurate boundary detection in high-resolution remote sensing images. UV-Mamba mitigates the memory loss problem in lengthy sequence modeling, which arises in state space models with increasing image size, by incorporating deformable convolutions. Its architecture utilizes an encoder-decoder framework and includes an encoder with four deformable state space augmentation blocks for efficient multi-level semantic extraction and a decoder to integrate the extracted semantic information. We conducted experiments on two large datasets showing that UV-Mamba achieves state-of-the-art performance. Specifically, our model achieves 73.3\% and 78.1\% IoU on the Beijing and Xi'an datasets, respectively, representing improvements of 1.2\% and 3.4\% IoU over the previous best model while also being 6$\times$ faster in inference speed and 40$\times$ smaller in parameter count. Source code and pre-trained models are available at \url{https://github.com/Devin-Egber/UV-Mamba}.

% Source code and pre-trained models are available in the supplementary material.

\end{abstract}


\section{Introduction}
\begin{figure}[t]
  \centering
  \noindent\includegraphics[width=0.85\linewidth]{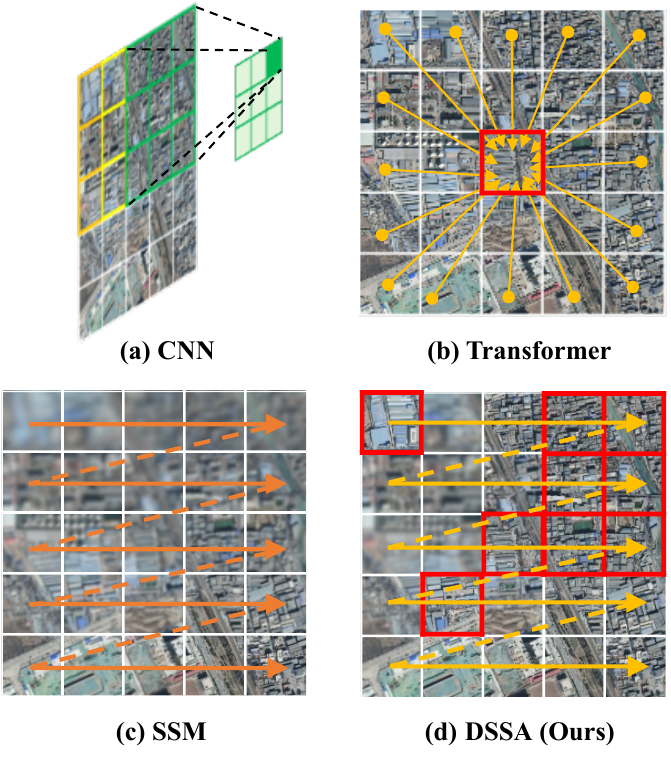}
  \caption{Comparisons of different core operators for establishing image patch correlation, with patch opacity representing the degree of memory loss. The local characteristics of (a) CNN and the quadratic complexity of (b) Transformer limit their ability to achieve fine-grained global modeling. (c) SSM has limited long-distance modeling capabilities, and (d) DSSA mitigates memory loss in SSM during long sequence modeling.}
  \label{fig:compare}
\end{figure}

\begin{figure*}
  \centering
  \noindent\includegraphics[width=0.98\linewidth]{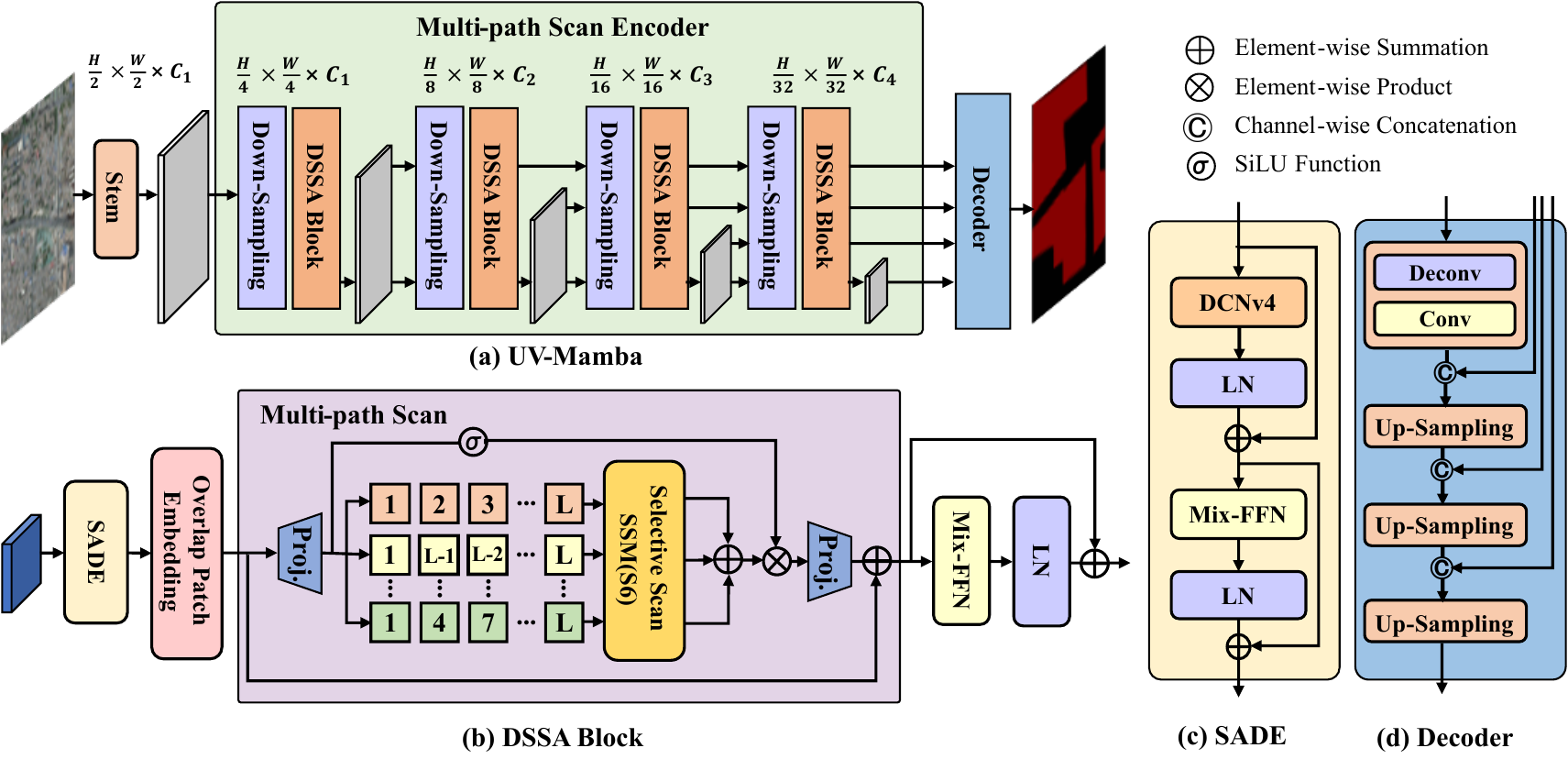}
  \caption{Overview architecture of our proposed UV-Mamba.}
  \label{fig:overall_architecture_diagram}
\end{figure*}

As historical remnants in the urbanization process, urban villages present significant urban planning and management challenges because of their low-rise and densely packed buildings, substandard environmental conditions, and outdated municipal infrastructure~\cite{urbanization, redevelopment, development, optimizing}. The issue of urban villages not only concerns the aesthetic and cleanliness of the city's image but also directly affects residents' quality of life, public safety, and social stability~\cite{village, urbanvalues, housing}. Traditional methods of collecting information on urban villages mainly rely on manual field surveys, which are time-consuming and labor-intensive~\cite{zheng2009urban}. 

To automatically identify urban village boundaries, exploring image segmentation techniques using satellite imagery has garnered widespread attention~\cite{gaofen, unsupervised, finegrained, Pansharpen}. Several studies have employed advanced semantic segmentation models, including Fully Convolutional Networks and U-Net, to map urban village areas~\cite{urbanFCN, urbanU-Net}. ~\cite{DomainAdaption} utilizes adversarial learning to fine-tune the semantic segmentation network, thereby adaptively generating consistent outputs for input images across various domains. UisNet~\cite{UisNet} enhances segmentation accuracy by integrating features from remote sensing imagery and building contours through a spatial-channel feature fusion module. UV-SAM~\cite{UV-SAM} capitalizes on the strengths of both a general model and a specialized model to apply the zero-shot capabilities of SAM~\cite{SAM} to the task of urban village boundary identification.

However, accurately delineating the boundaries of urban villages in existing research is challenging due to two primary factors. First, the unique architectural characteristics of urban villages, including high density, narrow streets, and diverse building forms, pose inherent difficulties. Second, the limitations of CNN in capturing global information and the computational complexity of transformers~\cite{dcnv3, SPMamba, zou2023pmaa}, as shown in Fig. \ref{fig:compare}, further complicate this task. Moreover, spatial features and dependencies can be lost when ultra-high-resolution (UHR) remote sensing images are divided into smaller patches. 

To address the above issues, we propose the UV-Mamba model, which leverages the global modeling capability of SSM with linear complexity and deformable convolutions' spatial geometric deformation ability. Our model mitigates the memory loss issue of SSM in lengthy sequence modeling by employing DCN to allocate greater weights to regions of interest, thereby improving SSM's capacity to retain information across extended sequences. The main contributions of our architecture are summarized as follows:

\begin{itemize}
\item We introduce UV-Mamba, a novel and efficient architecture based on SSM that effectively preserves linear computational complexity while delivering enhanced global modeling capabilities.

\item We design a DSSA module that mitigates memory loss in SSM during long-distance modeling as the sequence grows by assigning greater weights to regions of interest using deformable convolutions.

\item We conduct extensive experiments on two cities, Beijing and Xi’an, in China. The results show that our method achieves superior performance, surpassing the state-of-the-art CNN-based and Transformer-based models.

\end{itemize}

\section{METHODOLOGY}

\subsection{Preliminaries: State Space Model}

The state space model derives from the linear time-invariant systems in modern control theory. It maps a one-dimensional input signal $X(t) \in  \mathbb{R}$ to an N-dimensional latent state $h(t) \in  \mathbb{R}^{N}$, and then projects it to a one-dimensional output signal $y(t)$. This process can be described by the following linear ordinary differential equations (ODE):
\begin{equation}
\begin{aligned}
h'(t) &= \mathbf{A}h(t) + \mathbf{B}x(t), \\
y(t) &= \mathbf{C}h(t),
\end{aligned}
\label{equ01}
\end{equation}
where $\mathbf{A} \in  \mathbb{R}^{N \times N}$ is the state transition matrix, $\mathbf{B} \in  \mathbb{R}^{N}$ and $\mathbf{C} \in  \mathbb{R}^{N}$ are the projection matrix.

To better adapt to the discrete inputs in deep learning such as text sequences, $\mathbf{A}$ and $\mathbf{B}$ are discretized using a zero-order hold (ZOH) technique with a learnable time scale parameter $\boldsymbol{\Delta}$, which transforms the continuous SSM into a discrete SSM. The process is as follows:

\begin{equation}
\begin{aligned}
\overline{\mathbf{A}} &= exp(\boldsymbol{\Delta} \mathbf{A}), \\
\overline{\mathbf{B}} &= (\boldsymbol{\Delta}\mathbf{A})^{-1}exp(\boldsymbol{\Delta} \mathbf{A} - \mathbf{I}) \cdot \boldsymbol{\Delta}\mathbf{B}.
\end{aligned}
\end{equation}

After discretization, the Eq. \ref{equ01} can be represented as:
\begin{equation}
\begin{aligned}
{h}_{k} &= \overline{\mathbf{A}}{h}_{k-1} + \overline{\mathbf{B}}{x}_{k}, \\
{y}_{k} &= \overline{\mathbf{C}}{h}_{k},
\end{aligned}
\label{equ03}
\end{equation}
where $\overline{\mathbf{A}}$ and $\overline{\mathbf{B}}$ represent the discretized versions of the $\mathbf{A}$ and $\mathbf{B}$ matrices, respectively. ${h}_{k-1}$ represents the previous state information and ${h}_{k}$ represents the current state information.

% ${h}_{k-1}$ represents the state information from the previous time step, and ${h}_{k}$ represents the current state information.

\begin{figure}[t]
  \centering
  \noindent\includegraphics[width=0.98\linewidth]{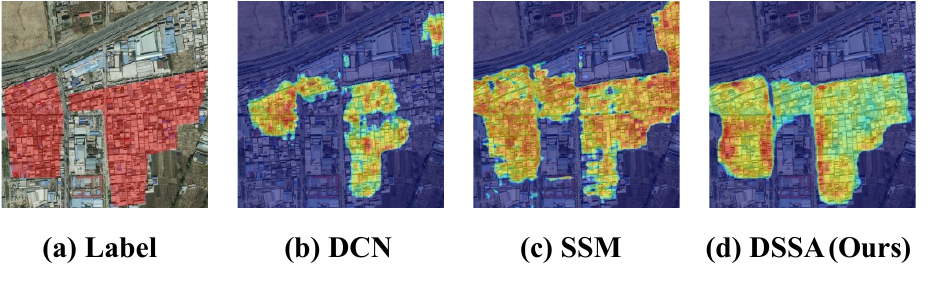}
  \caption{Comparison of class activation map across operators.}
  \label{fig:cam_map}
\end{figure}

\subsection{Architecture Overview}

The architecture of the proposed UV-Mamba model, as depicted in Fig. \ref{fig:overall_architecture_diagram} (a), is composed of three principal components: 
A stem module with varying convolutional kernel sizes, a hierarchical multi-path scan encoder, and a lightweight decoder. The stem module, which performs initial feature extraction and downsamples the input image by a factor of 2, consists of four convolutional layers with 7 $\times$ 7 and 3 $\times$ 3 kernels, padding of 3 and 1, and strides of 2 and 1, respectively. The multi-path scan encoder consists of four deformable state space augmentation (DSSA) blocks, which progressively reduce the feature map size by half at each stage, resulting in feature maps of various scales relative to the model input: $\frac{H}{4} \times \frac{W}{4} \times {C}_{1}$, $\frac{H}{8} \times \frac{W}{8} \times {C}_{2}$, $\frac{H}{16} \times \frac{W}{16} \times {C}_{3}$, $\frac{H}{32} \times \frac{W}{32} \times {C}_{4}$. The decoder comprises four upsample modules, each incorporating a transposed convolution to upsample the feature map from the encoder by a factor of two, followed by two 3 $\times$ 3 convolutions for feature fusion. Finally, bilinear interpolation is used to restore the image to the input size.

\subsection{Deformable State Space Augmentation Block}

Two primary challenges for UHR remote sensing of dense urban environments are refining pixel-level representation and ensuring robust global modeling for accurate boundary extraction. To address these challenges, we design the DSSA Block, which includes patch embeddings, a spatially adaptive deformable enhancer (SADE), a multi-path scan SSM module (MSSM), and patch merging, as illustrated in Fig. \ref{fig:overall_architecture_diagram} (b). Notably, our SADE and MSSM modules are stacked twice as intermediate modules. The issue of memory loss during global modeling with SSM can be mitigated by assigning greater weights to regions of interest through the SADE. This approach achieves linear complexity while enhancing global modeling capabilities beyond SSM, enabling more effective differentiation between buildings, as shown in Fig. \ref{fig:cam_map}.

\textbf{Multi-path Scan SSM Module.} 
A series of studies~\cite{mamba, VisionMamba, VMamba, Rs-mamba} have demonstrated that increasing the number of scanning directions in SSM-based models is crucial for achieving comprehensive global modeling capabilities. To better delineate the boundaries between urban villages and adjacent communities, we aggregate scanning results from eight directions (horizontal, vertical, diagonal, and anti-diagonal, both forward and backward) to capture the complex spatial relationships of surrounding structures and to provide a thorough understanding of the contextual environment. To better adapt to varying input sizes, we introduce Mix-FFN~\cite{leformer}, which is more effective in providing positional information than traditional positional encoding~\cite{PositionalEncodings, SegFormer, PositionInformation,chen2023high}, by applying a 3 $ \times$ 3 convolution within the feed-forward network.

\textbf{Spatially Adaptive Deformable Enhancer.} As shown in Fig. \ref{fig:overall_architecture_diagram} (c), the design of the SADE adopts a structure similar to that of the transformer~\cite{Transformer}. By leveraging the spatial geometric deformation learning capabilities of the deformable convolution~\cite{dcn}, it more effectively adapts to urban villages' diverse spatial distribution characteristics. Specifically, we utilize the DCNv4~\cite{DCNV4} operator for spatial feature enhancement, valued for its fast convergence and processing efficiency. The process is as follows:
\begin{equation}
\begin{aligned}
\mathbf{y}({p}_{0}) = \displaystyle\sum_{g=1}^{G}\displaystyle\sum_{k=1}^{K}{\mathbf{w}}_{g}{\mathbf{m}}_{gk}{\mathbf{x}}_{g}({p}_{0} + {p}_{k} + \Delta{p}_{gk} ),
\end{aligned}
\end{equation}
\\
where $G$ denotes the total number of aggregation groups. 
For the g-th group, ${w}_{g}$ represents the location-irrelevant projection weights, ${m}_{gk}$ is the modulation scalar for the k-th sampling point, ${x}_{g}$ denotes the sliced input feature map, and $\Delta {p}_{gk}$ is the offset for the grid sampling location $\Delta {p}_{k}$. The extracted features are then further aggregated using Mix-FFN, which reduces computational complexity while maintaining the model's representational capacity.

% The extracted features are aggregated using Mix-FFN, reducing computational complexity while maintaining the model's representational capacity. 

\begin{figure*}[t]
  \centering
  \noindent\includegraphics[width=0.98\linewidth]{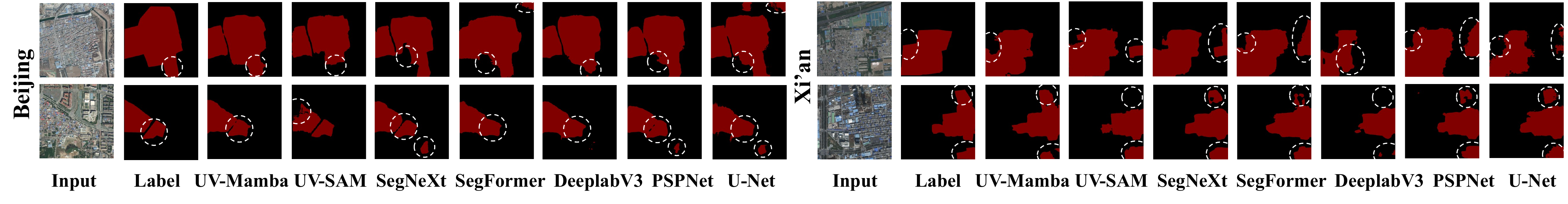}
  % \noindent\includegraphics{image/pred_mask.pdf}
  \caption{Visualization results of our proposed UV-Mamba and other methods. The white circles indicate apparent differences.}
  \label{fig:control_experiment}
\end{figure*}

\section{Experiments}

\subsection{Experimental Settings}

\textbf{Dataset.} 
We use datasets from Beijing and Xi'an~\cite{UV-SAM}, two Chinese cities with distinct architectural styles due to their significant geographical differences. Both cities feature a mix of traditional and modern buildings, creating complex urban structures that challenge our model in extracting urban village boundaries. The Beijing dataset contains 531 images, while the Xi'an dataset comprises 205 images. We divided these datasets into training, validation, and test sets in a 6:2:2 ratio. Each image has a resolution of 1024 $\times$ 1024 to ensure the inclusion of the primary urban information.  

\textbf{Implementation Detail.} Our experiments are conducted on a single Tesla V100 GPU, training for 100 epochs. To prevent overfitting and improve generalization, we applied a consistent data augmentation strategy across all experiments, which included random rotation, horizontal flipping, and vertical flipping. The model is initially pre-trained on the Cityscapes dataset~\cite{Cityscapes} and subsequently fine-tuned on the urban village dataset. We use the Adam~\cite{adamW} optimizer during pre-training with an initial learning rate of 0.001. The learning rate is warmed for the first 10 epochs and decreases gradually to 1e-6. Cross Entropy loss~\cite{cross-entropy} is utilized during pre-training to optimize the model's performance.

% Cross Entropy loss~\cite{cross-entropy} is employed for the pre-training phase.

The pre-trained weights are then fine-tuned on the Urban Village dataset. To fine-tune the Beijing and Xi'an datasets, we continue to use the Adam optimizer. The learning rate is warmed up for the first 30 epochs and gradually decreases to 1e-6. Specifically, we set the learning rate for the Beijing dataset to 0.0004 and use the Dice loss function~\cite{DiceLoss}. For the Xi'an dataset, the learning rate is set to 0.0002, and the Cross Entropy loss function is employed. The models' accuracy is evaluated using Intersection over Union (IoU), accuracy (ACC), and overall accuracy (OA). The efficiency is assessed by the parameter (Params, M) and floating point operations per second (Flops, G), denoted as \#P and \#F, respectively, in the table for brevity.

\subsection{Ablation Studies}

\begin{table}[t]
% \tiny
% \small
\centering
\caption{Performance of UV-Mamba on Beijing and Xi'an dataset with different image sizes.}
\fontsize{9}{10}\selectfont
\setlength{\tabcolsep}{5pt}
\begin{tabular}{c|ccc|ccc}
\toprule%第一道横线
\multirow{2}{*}{Image Size}  & \multicolumn{3}{c|}{Beijing}    & \multicolumn{3}{c}{Xi'an}          \\ \cline{2-7} 
                                                                      & IoU$\uparrow$                              & ACC$\uparrow$                        & OA$\uparrow$                            & IoU$\uparrow$                        & ACC$\uparrow$                        & OA$\uparrow$                            \\ \hline
128                & 0.631 & 0.762 & 0.938 & 0.541 & 0.597 & 0.920 \\
256                & 0.645 & 0.782 & 0.940 & 0.637 & 0.680 & 0.939 \\
512                & 0.679 & 0.787 & 0.948 & 0.681 & 0.720 & 0.947 \\
1024               & \textbf{0.733}  & \textbf{0.837}   & \textbf{0.957}   & \textbf{0.781}  & \textbf{0.853}  & \textbf{0.962}                  \\ 

\bottomrule%第四道横线
\end{tabular}
\label{table:ablation_study_img_size}
\end{table}

\begin{table}[t]
% \scriptsize
% \small
\centering
\caption{Impact of different encoder combinations on \#P and \#F. P, R, and S, respectively, represent whether the positions of the two modules are parallel, reverse, or serial.
}
\setlength{\tabcolsep}{3.5pt}
\fontsize{7.8}{10}\selectfont
\begin{tabular}{c|cc|ccc|c|c|c|c|c}
\toprule%第一道横线
\multirow{2}{*}{Dataset} & \multicolumn{2}{c|}{DSSA}     & \multicolumn{3}{c|}{Position}     & \multirow{2}{*}{\#P$\downarrow$} & \multirow{2}{*}{\#F$\downarrow$} & \multirow{2}{*}{IoU$\uparrow$}              & \multirow{2}{*}{ACC$\uparrow$}        & \multirow{2}{*}{OA$\uparrow$}            \\ \cline{2-6}
                         & SADE           & MSSM         & P   & R   & S      &                            &                           &                                  &                                  &                                  \\ \hline
\multirow{6}{*}{Beijing}      & \XSolidBrush & \Checkmark & \XSolidBrush   & \XSolidBrush  & \Checkmark       & 5.7    & 139.7           & 0.709  & 0.829  & 0.952  \\
                         & \Checkmark   & \XSolidBrush   & \XSolidBrush    & \XSolidBrush  & \Checkmark       & 3.8    & 124.6           & 0.705  & 0.820  & 0.952 \\
                         & \Checkmark   & \Checkmark     & \Checkmark      & \XSolidBrush  & \XSolidBrush     & 7.7    & 153.7           & 0.727  & 0.835  & 0.956 \\
                         & \Checkmark   & \Checkmark     & \XSolidBrush    & \Checkmark    & \XSolidBrush     & 7.7    & 153.6           & 0.708  & 0.832  & 0.952 \\
                         & \Checkmark   & \Checkmark     & \XSolidBrush   & \XSolidBrush   & \Checkmark       & 7.7    & 153.6           & \textbf{0.733}  & \textbf{0.837}  & \textbf{0.957} \\
\hline                        
\multirow{6}{*}{Xi'an}   & \XSolidBrush & \Checkmark & \XSolidBrush   & \XSolidBrush  & \Checkmark            & 5.7    & 139.7           & 0.726 & 0.777 & 0.954 \\
                         & \Checkmark   & \XSolidBrush   & \XSolidBrush    & \XSolidBrush  & \Checkmark       & 3.8    & 124.6           & 0.714 & 0.772 & 0.951 \\
                         & \Checkmark   & \Checkmark     & \Checkmark      & \XSolidBrush  & \XSolidBrush     & 7.7    & 153.7           & 0.749 & 0.846 & 0.956 \\
                         & \Checkmark   & \Checkmark     & \XSolidBrush    & \Checkmark    & \XSolidBrush     & 7.7    & 153.6           & 0.704 & 0.811 & 0.946 \\
                         & \Checkmark   & \Checkmark     & \XSolidBrush   & \XSolidBrush   & \Checkmark       & 7.7    & 153.6           & \textbf{0.781} & \textbf{0.853} & \textbf{0.962} \\
\bottomrule%第四道横线
\end{tabular}
\label{table:ablation_study_b}
\end{table}

\textbf{Image Size:} To assess the impact of contextual information and spatial features on urban village boundary detection, we evaluate the model's performance using input images of varying sizes, with the results presented in Table \ref{table:ablation_study_img_size}. The results demonstrate that as image size increases, the accuracy of urban village detection consistently improves, likely due to the continuous spatial distribution of these areas. This finding highlights the importance of utilizing UHR remote sensing images for precise boundary detection.

\textbf{DSSA Module:} To evaluate the effectiveness of the DSSA module in UV-Mamba, we present the segmentation performance of different model variants on Beijing and Xi'an datasets in Table \ref{table:ablation_study_b}. The results indicate that the model's performance decreases by 2.4\% and 5.5\% without the SADE module. Similarly, without the MSSM module, performance drops by 2.8\% and 6.7\%. These findings underscore the importance of robust global modeling capabilities for accurate urban village segmentation. Furthermore, we experiment with various positional combinations of the SADE and MSSM modules within the DSSA module. The results showed that when the SADE and MSSM modules are arranged in parallel, the performance is suboptimal, achieving 72.7\% and 74.9\% IoU, respectively. Conversely, placing the SADE module after the MSSM module results in the worst overall model performance, suggesting that the long sequence modeling limitations of the SSM lead to feature map information loss, thereby misleading the model. In summary, these results indicate that the SADE can partially complement the global modeling capabilities of the SSM, helping to mitigate the memory loss issue when modeling high-resolution remote sensing images with the SSM.

\begin{table}[t]
\centering
\fontsize{8.8}{11}\selectfont
\caption{Quantitative comparison of our UV-Mamba and other methods on the Beijing and Xi'an datasets.}
\setlength{\tabcolsep}{2.1pt}
\begin{tabular}{c|c|ccc}
% \hline
\toprule%第一道横线
\multirow{2}{*}{Method}     & \multirow{2}{*}{\#P$\downarrow$}   & \multicolumn{3}{c}{Beijing / Xi'an}        \\            
\cline{3-5}                                               &                                             & IoU$\uparrow$         & ACC$\uparrow$           & OA$\uparrow$ \\    

\hline
                                      
U-Net~\cite{Unet}                                                    & 17.3                                        & 0.658 / 0.661          & 0.759 / 0.698          & 0.945 / 0.943            \\
PSPNet~\cite{PSPnet}                                                   & 46.6                                        & 0.697 / 0.759          & 0.834 / 0.842          & 0.949 / 0.958            \\
Deeplabv3~\cite{Deeplabv3}                                                & 39.6                                         & 0.706 / 0.592          & 0.832 / 0.664          & 0.951 / 0.927             \\
SegFormer~\cite{SegFormer}                                                & 13.7                                          & 0.699 / 0.713          & 0.821 / 0.794          & 0.951 / 0.949             \\
SegNeXt~\cite{SegNeXt}                                                  & 13.9                                        & 0.707 / 0.759          & 0.833 / 0.826          & 0.951 / 0.959              \\
UV-SAM~\cite{UV-SAM}                                                   & 316.2                                       & 0.721 / 0.747          & 0.807 / 0.804          & 0.953 / 0.957               \\
\textbf{UV-Mamba (Ours)}                                 & \textbf{7.7}                       & \textbf{0.733 / 0.781} & \textbf{0.837 / 0.853} & \textbf{0.958 / 0.962}  \\ 
\bottomrule%第四道横线
\end{tabular}
\label{table:results_uv}
\end{table}

\subsection{Comparison to the State-of-the-Arts}

As illustrated in Table \ref{table:results_uv}, UV-Mamba outperforms the advanced urban village identification models~\cite{Unet, PSPnet, Deeplabv3, SegFormer, SegNeXt, UV-SAM}, achieving the state-of-art performance on both the Beijing and Xi'an datasets. Fig. \ref{fig:control_experiment} presents the visualized segmentation results. Regarding segmentation accuracy, our model demonstrates a 1\%-3\% improvement in IoU across the two datasets compared to the previous best urban village boundary identification model UV-SAM with a parameter size that is 40 $\times$ smaller. Similar enhancements in performance are observed in the accuracy metrics of ACC and OA.

\section{Conclusion}

In this paper, we introduce the UV-Mamba model, which mitigates memory loss in long sequence SSM modeling, maintaining global modeling capabilities with linear complexity for precise segmentation and localization of urban village buildings in dense environments. We anticipate that our research will provide significant technical support for modernizing urban villages, thereby advancing urban development toward increased livability, harmony, and sustainability.

\bibliographystyle{IEEEtran}
\bibliography{strings}

\end{document}